\documentclass{article} % For LaTeX2e
\usepackage{iclr2025_conference,times}

\usepackage{amsmath,amsfonts,bm}

\def\eqref#1{equation~\ref{#1}}
\def\1{\bm{1}}

\DeclareMathAlphabet{\mathsfit}{\encodingdefault}{\sfdefault}{m}{sl}
\SetMathAlphabet{\mathsfit}{bold}{\encodingdefault}{\sfdefault}{bx}{n}

\usepackage{hyperref}
\usepackage{graphicx}
\usepackage{booktabs}
\usepackage{url}
\usepackage{wrapfig}
\usepackage{cancel}
\usepackage{amsmath,amsfonts,amssymb,amsthm}

\usepackage{listings}
\title{VLM Q-Learning: Aligning Vision-Language Models for Interactive Decision-Making}

\author{
Jake Grigsby$^{1}$\thanks{Work done during an internship at Salesforce AI Research.} , Yuke Zhu$^{1}$, Michael Ryoo$^{2}$, Juan Carlos Niebles$^{2}$\\
$^1$The University of Texas at Austin \quad
$^2$Salesforce AI Research\\
}

\iclrfinalcopy % Uncomment for camera-ready version, but NOT for submission.
\begin{document}

\maketitle

\begin{abstract}
Recent research looks to harness the general knowledge and reasoning of large language models (LLMs) into agents that accomplish user-specified goals in interactive environments. Vision-language models (VLMs) extend LLMs to multi-modal data and provide agents with the visual reasoning necessary for new applications in areas such as computer automation. However, agent tasks emphasize skills where accessible open-weight VLMs lag behind their LLM equivalents. For example, VLMs are less capable of following an environment's strict output syntax requirements and are more focused on open-ended question answering. Overcoming these limitations requires supervised fine-tuning (SFT) on task-specific expert demonstrations. Our work approaches these challenges from an offline-to-online reinforcement learning (RL) perspective. RL lets us fine-tune VLMs to agent tasks while learning from the unsuccessful decisions of our own model or more capable (larger) models. We explore an off-policy RL solution that retains the stability and simplicity of the widely used SFT workflow while allowing our agent to self-improve and learn from low-quality datasets. We demonstrate this technique with two open-weight VLMs across three multi-modal agent domains.
\end{abstract}
\section{Introduction}

\label{sec:introduction}

Diverse training datasets give large language models (LLMs) the ability to generate knowledgeable text across a wide range of topics. LLM pre-training produces a model that outputs the most likely continuation of its input text according to a distribution of training examples sourced from across the web. However, there are many applications where we would like to direct our model to accomplish a specific objective. In these tasks, our model acts as an agent in an interactive environment and makes a sequence of decisions to maximize an evaluation metric. Because the model was not trained to optimize this objective, the best decision-making strategy may be an unlikely output. Ongoing research looks to fix this alignment problem and repurpose generative models for decision-making. 

LLM-agent techniques are largely enabled by the post-training process, where the model is fine-tuned on curated datasets of chat or assistant-like behavior. A natural first step is to continue training with supervised fine-tuning (SFT) on task-specific datasets. However, SFT cannot outperform the best decision-making strategy in its training data and relies on collecting successful demonstrations. LLMs' ability to adapt to new tasks based on examples in their input \citep{gpt3} allows this realignment to be done without further training: agent behavior can be generated by prompting an LLM with demonstrations. Additional prompting techniques improve long-term planning by eliciting self-reflection and reasoning \citep{yao2022react, wei2022chain}.

Vision-language models (VLMs) extend LLMs to multi-modal inputs. VLMs are trained by using smaller datasets of interleaved text and image data to mix visual features into an existing LLM \citep{liu2024visual, liu2024improved}. The high-level problem of realigning VLMs with decision-making objectives is similar to that of LLM agents. However, VLM research is still in its early stages, and current models lag behind their LLM equivalents in two critical areas: 1) agent action syntax and 2) long-context prompting. Agent tasks often do not involve freeform dialogue and instead require outputs with a specific syntax that the environment can interpret. The ability to plan and act in this more restricted output space is closely related to the tool-use (or function-calling) capabilities of LLMs \citep{qu2024tool}. Perhaps a more long-term issue is that environment interactions can stretch over many model outputs, making it impractical to include complete in-context examples and align the VLM via prompting. SFT can realign a base model with agent tasks, but it is only effective when we can demonstrate a strategy that meets our performance expectations. 
\begin{wrapfigure}{r}{.55\textwidth}
\begin{center}
    \includegraphics[width=.54\textwidth]{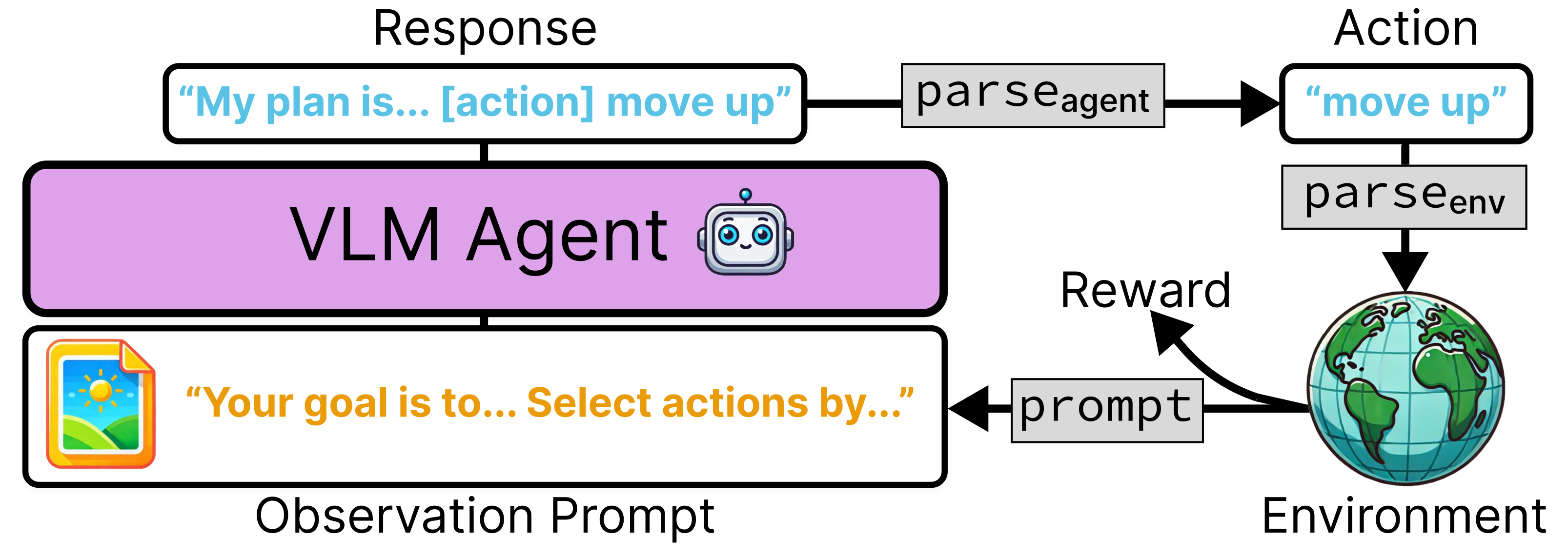}
    \vspace{-2mm}
    \caption{\textbf{VLM-Environment Interaction.} We format text and image data from the environment into an input prompt for the VLM. The models's text reply is parsed into a concrete action executed by the environment to produce a reward and new inputs for the VLM.}
    \label{fig:fig1}
    \vspace{-4mm}
\end{center}
\end{wrapfigure}

Our work approaches the challenges of fine-tuning VLMs for agent tasks from a reinforcement learning (RL) perspective (Figure \ref{fig:fig1}). RL solves the agent alignment issue by explicitly training our model to maximize success in its environment. Decision-making is often formalized as an RL problem, but RL methods have a reputation for adding complexity; simpler prompting and SFT techniques are effective for LLM-agent tasks and are more widely used in recent work. However, we note that a category of offline RL techniques \citep{levine2020offline} that filter demonstration datasets according to a learned value model \citep{wang2020critic} have important advantages in the emerging area of VLM agent training. Like most offline RL algorithms, these filtered supervised fine-tuning (FSFT) methods let our agent outperform its demonstrations. Unlike other offline techniques, FSFT is effectively identical to SFT when the dataset is of high quality and lets us seamlessly collect new data when our agent needs further improvement. This lets FSFT serve as a low-downside, high-upside replacement for SFT in VLM agent domains where SFT is widely used. Our work explores the challenges of applying VLMs to agent tasks and demonstrates the strengths of off-policy RL applied to two open-weight models across three domains.

\vspace{-5mm}
\section{Background}
\vspace{-2mm}
\label{sec:background}

\noindent\textbf{Vision-Language Models.} VLMs extend the capabilities of LLMs to visual data by taking text and images as input and producing text outputs. They process multi-modal inputs by projecting the features of a vision encoder into the embedding space of an LLM — allowing the LLM to attend to the features of both modalities. While there have been many VLM models released in recent months, our work will focus on publicly available (open-weight) models that are small enough to be fine-tuned with accessible GPU resources. Table \ref{tbl:vlm_intro} introduces three models used in our experiments: PaliGemma \citep{Beyer2024paligemma}, xGen-MM \citep{xue2024xgen}, and MoonDream2 \citep{vik_2024}.

\begin{wraptable}{r}{.44\textwidth}
\vspace{-4mm}
\centering
\resizebox{.43\textwidth}{!}{%
\begin{tabular}{@{}lccc@{}}
\toprule
\multicolumn{1}{c}{} & 
\textbf{MoonDream2} & \textbf{PaliGemma} & \textbf{xGen-MM} \\ \midrule
Parameters     & 1.9B       & 2.9B      & 4.6B    \\
Feature Dim.         & 2048          & 2048         &  3072       \\
Vocabulary      & 51,200        & 257,216      & 32,015     \\ \bottomrule
\end{tabular}
}
\vspace{-2mm}
\caption{\textbf{VLMs in Our Experiments.}}.
\label{tbl:vlm_intro}
\vspace{-6mm}
\end{wraptable}

\noindent\textbf{Foundation Models as Agents.} We focus on agent applications where a model is evaluated based on its ability to make decisions that achieve a specific outcome. Examples include: controlling an operating system with text commands  \citep{wu2024oscopilot}, writing code to answer data analysis questions \citep{hu2024infiagent}, navigating graphical interfaces, \citep{gao2023assistgui, tao2023webwise, xie2023openagents}, ordering items from a web store \citep{yao2022webshop, tao2023webwise}, carrying out administrative tasks in business software \citep{drouin2024workarena}, making PowerPoint slides based on a specific topic \citep{guo2023pptc}, and playing games \citep{chen2024llmarena}, among many others. These tasks are characterized by \textit{multi-turn interactions}, where the model iteratively reacts to changes in the environment until it is successful. Success is measured by explicit task-specific objectives and not by the likelihood of the model's response or subjective human preferences. This creates a fundamental mismatch between the objectives the model was trained to maximize and the objective we are evaluating.

We can realign foundation models with agent tasks by biasing their outputs with careful prompting. Agent tasks require high-level planning over a sequence of outcomes. Prompting techniques for agent applications combine in-context learning \citep{gpt3} with step-by-step reasoning and self-reflection over previous outcomes \citep{wei2022chain, yao2022react, yang2023auto, zheng2024gpt}. Prompting is particularly effective when working with large proprietary models that support sequence lengths long enough to let these methods grow arbitrarily complicated by making multiple API calls to correct mistakes, retrieve relevant information, and plan for the future \citep{topsakal2023creating, xiao2023o3d, lutz2024wilbur, sridhar2023hierarchical}.

\noindent\textbf{Tool Use and Function Calling.} In order for an external environment like a browser or game to carry out an agent's request, it must be able to interpret the model's intention from open-ended text. Therefore, our model needs to generate text according to a task-specific syntax that is described as part of its input prompt \citep{hsieh2023tool}. Action syntax creates a barrier to entry where models fail because they cannot produce valid outputs despite responses that suggest they have some understanding of the task. This problem is closely related to the ``tool-use'' (or function-calling) capabilities of LLMs \citep{ling2023international, xu2023tool}. Tool-use refers to the ability of an LLM to call external APIs and is an important feature of chat assistants that retrieve information on current events, perform calculations, or prompt specialized image generation models. Tool-use accuracy in accessible open-weight LLMs once lagged behind proprietary models but has been improved by fine-tuning on specialized datasets with standardized syntax \citep{patil2023gorilla, qin2023toolllm}.

\noindent\textbf{Fine-Tuning Agents.} The syntax of agent environments is less standardized than tool-use, and task inputs may be very different from those seen during training. When prompting alone is not enough to create a successful agent, performance can be improved by fine-tuning on task-specific demonstrations \citep{zeng2023agenttuning, hong2024cogagent, feng2023llama, song2024trial, zhang2024agentohana, yin2023lumos}. Continuing training can be expensive, but we can safely assume that the resulting agent will match the performance of the demonstrator(s) when given sufficient training data. However, we cannot expect our agent to learn a better strategy than appeared in the fine-tuning demonstrations because it is trained to replicate the decisions in the dataset.

\noindent\textbf{Off-Policy RL.} 
RL formalizes the process of learning optimal decision-making strategies in interactive environments. The environment begins in a state $s_0$ that is represented to the agent by an observation $o_0$. The agent's policy $\pi$ takes this observation as input and samples an action $a_0 \sim \pi(o_0)$ from the space of possible actions $\mathcal{A}$, leading to a new state $s_1$ and reward $r_1$ determined by the environment's transition function ($T(s_0, a_0) = s_1$) and reward function ($R(s_0, a_0) = r_1$), respectively. This process repeats after the agent receives its next observation $o_1$. RL uses data to discover a policy that maximizes the agent's return --- or cumulative reward from the current state: $V^{\pi}(s_t) = \mathop{\mathbb{E}}_{\pi}\left[\sum_{i=t}^{\infty}\gamma^{i-t}r_i\right]$, where $\gamma \in [0, 1)$ is a discount factor.

\textit{Off-policy} RL methods can learn from any experience in the environment --- regardless of whether the actions of the current policy generated that experience. An example is $Q$-Learning \citep{watkins1992q, mnih2015human}, which learns a function $Q$ to estimate the return that will follow a particular action $a$: $Q^{\pi}(s_t, a_t) = R(s_t, a_t) + V^{\pi}(s_{t+1})$. If we can learn an accurate estimate of $Q$, we can improve the policy by selecting actions that maximize $Q$. Actor-critic methods train a critic model to approximate $Q$ and a second (often entirely separate) actor model to maximize the critic's predictions \citep{lillicrap2015continuous}. \textit{Offline} RL agents learn a policy from a fixed dataset \citep{levine2020offline}. It is common to view offline RL as an extreme case of off-policy RL where we assume that our dataset was generated by a mixture of unknown policies and that we cannot collect additional data. However, vanilla $Q$-Learning methods struggle to handle the fully offline setting because they repeatedly maximize their current $Q$-values but cannot collect new experiences that would reveal their predictions are overestimated \citep{kumar2019stabilizing}. Successful modifications avoid directly maximizing $Q$ \citep{kostrikov2021offline} or regularize estimates to be more conservative \citep{kumar2020conservative, agarwal2020optimistic}. We derive our method from a family of filtered (or weighted) behavior cloning algorithms where a value estimate is used to mask (or downweight) sub-optimal demonstrations \citep{nair2020awac, peng2019advantageweightedregressionsimplescalable, nair2018overcoming, chen2020bail, wang2018exponentially}; the variant used here will be most similar to CRR \citep{wang2020critic}. These methods are capable of outperforming the quality of their dataset by learning to prefer alternative actions when they lead to higher $Q$-values \citep{kumar2022should, ghugare2024closing}. \textit{Online} RL refers to the more standard setting where we are allowed to interact with the environment to collect additional data. \textit{On-policy} updates improve a policy based on its own recent experience \citep{williams1992simple, schulman2015trust}, and are generally less sample efficient.

\vspace{-2mm}
\section{VLM Agents as RL Policies}
\label{sec:vlm_rl}

We approach agent interactions from an RL perspective, where the VLM will become an RL policy mapping text and image observations to text actions. Text is represented by tokens from the VLM's vocabulary, $\mathcal{V}$. The external task the agent interacts with --- such as the browser or game it is playing --- can be thought of as the environment. The environment interprets text actions to produce the next observation and reward. Because VLMs are sequence-to-sequence models, there are two ways to define the action space $\mathcal{A}$ \citep{zhouarcher, abdulhai2023lmrl}. One option is to view each individual output token as a separate action ($\mathcal{A} = \mathcal{V}$). This \textbf{``token-based view''} most closely aligns with how sequence models are trained. During evaluation and data collection, it is simpler to take a ``\textbf{turn-based view}'' where each timestep is a full turn of dialogue and actions are complete replies with some maximum length $r$ ($\mathcal{A} = \mathcal{V}^{r}$). On each turn $t$, the environment provides an observation of text ($o^{\text{text}}$) and image ($o^{\text{img}}$) data, and we query the VLM to generate a complete response $a_t$. The open-ended text response is parsed by the environment to produce the true action that it uses to update its state. Depending on the environment, the ``true'' action space $\mathcal{A}'$ may be text, code, or a set of discrete choices. We refer to this mapping from text actions to true actions as $\texttt{parse}_{\text{env}}: \mathcal{A} \rightarrow \mathcal{A}'$. We assume this behavior is outside the agent's control and that it is the agent's responsibility to produce text actions that $\texttt{parse}_{\text{env}}$ can interpret. Figure \ref{fig:fig1} summarizes the turn-based interaction between a VLM agent and its environment. Our goal is to use RL to update the VLM's parameters to maximize rewards.

\vspace{-2mm}
\section{Related Work}

RL has many applications in foundation model fine-tuning and agent tasks. A summary of related work in Reinforcement Learning from Human Feedback and applications of foundation models without open-ended text-based actions is deferred to Appendix \ref{app:additional_related}.

Our VLM agent method shares technical motivations with work on fine-tuning LLMs for multi-turn dialogue. WebRL \citep{qi2024webrl} applies a constrained on-policy RL update to prevent the fine-tuned policy from diverging from an initial task-specific SFT policy. When combined with a variety of other curriculum and dataset filtering techniques, the WebRL update allows open-source models to significantly outperform proprietary LLMs in  WebArena \citep{zhou2023webarena}. ILQL \citep{snelloffline} adjusts the outputs of a base LLM during inference based on the predictions of a second critic LLM trained by offline RL. ArCHer \citep{zhouarcher} is an online RL update that highlights the ``turn-based'' vs. ``token-based'' action discrepancy (Section \ref{sec:vlm_rl}) and creates a hierarchical approach that splits learning across both timescales. In contrast, we will train RL solely from the token-based format because it allows for a simple substitution of SFT. ILQL and ArCHer are actor-critic updates; like most actor-critics, they create training signal for the main policy model we are fine-tuning by training other models to learn value predictions ($Q^{\pi}$ and $V^{\pi}$) or a reference policy based on SFT. The cost of training multiple models is rarely an issue at the scale of standard RL research domains but is a significant concern in VLM applications where models contain billions of parameters. Our method will only involve a single VLM.

Recent work has applied online RL to VLMs in multi-turn vision tasks \citep{fereidouni2024search}. The most closely related work to our own is RL4VLM \citep{zhai2024fine}. RL4VLM fine-tunes the LLaVA VLM \citep{liu2024visual}  in a two-stage process. During the first stage, the base model is aligned to the syntax of the environment by SFT on demonstrations generated by chain-of-thought prompting \citep{wei2022chain}. The second stage improves the success rate of the SFT model with on-policy RL via PPO \citep{schulman2017proximal}. Our method will replace this two-stage pipeline with a single stage by using an RL update that makes the SFT stage redundant.

\vspace{-2mm}
\section{Method}
\label{sec:method}

Our method follows the popular framework of performing SFT on task-specific data that is generated by basic prompting. However, we allow the agent to improve over its dataset by filtering out tokens that we estimate will degrade the agent's performance. This filtering process is enabled by converting VLM agent fine-tuning to an off-policy actor-critic RL problem.

\subsection{Task Setup and Prompting}

\label{sec:prompting}

The environment provides an observation of text ($o^{\text{text}}$) and image ($o^{\text{img}}$) data. A \texttt{prompt} method reformats information from the environment into a full input prompt, including a description of the agent's objective. We include documentation of the actions available to the agent alongside an example of the syntax expected by $\texttt{parse}_{\text{env}}$ (Section \ref{sec:vlm_rl}). We can generalize our method by introducing a second parser: $\texttt{parse}_{\text{agent}}$ (Figure \ref{fig:fig1}). $\texttt{parse}_{\text{agent}}$ enables arbitrary output formats including chain-of-thought-style reasoning \citep{wei2022chain, yao2022react}. We will allow for reasoning (or ``thought'') outputs by setting $\texttt{parse}_{\text{agent}}$ to return only the text after the string ``\texttt{[action]}''. This choice is arbitrary and could be replaced by a JSON output (as in RL4VLM), for example. Our VLM agents' first challenge will be to output text actions $a$ that are valid inputs to $\texttt{parse}_{\text{env}}(\texttt{parse}_{\text{agent}}(a))$. When the agent fails to do so, it will be given an error message and prompted to try again. VLMs that cannot break out of this loop create a sparse reward problem by failing to reach useful reward signals for training. We minimize this problem by introducing penalties for invalid syntax on top of the environment's native reward function (Appendix \ref{app:rewards}). Appendix \ref{app:action_prompting} provides further discussion on action parsing and its connection to RL exploration.

\subsection{RL Fine-Tuning}
\label{sec:vlmq}

\vspace{-2mm}

\begin{figure}[h!]
    \centering
    \includegraphics[width=0.95\linewidth]{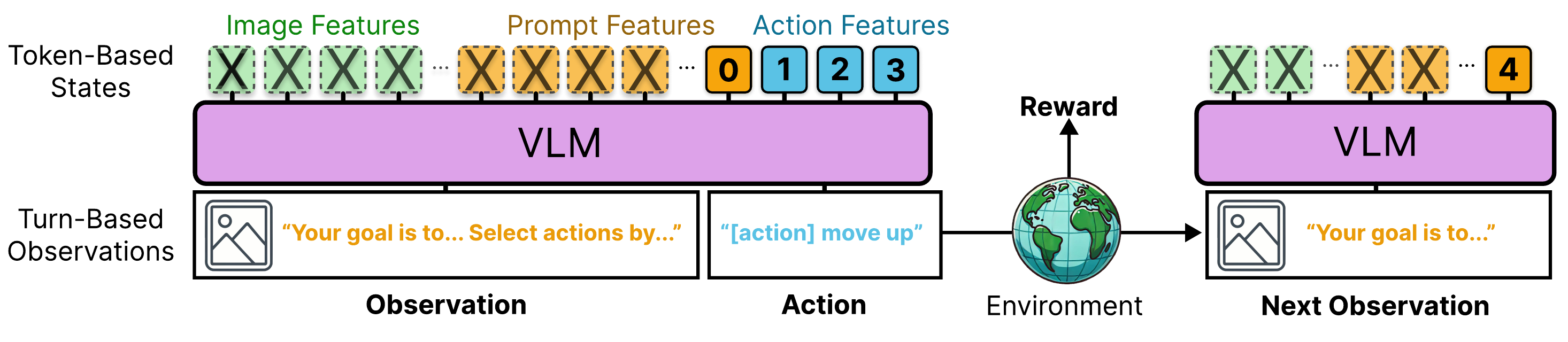}
    \vspace{-2mm}
    \caption{\textbf{Converting Turns $\rightarrow$ Tokens.} We treat each token of the agent's reply as a single action by passing consecutive turns of dialogue through the VLM and finding output representations corresponding to tokens where an action decision is made. Outputs corresponding to the prompt tokens are masked and ignored. The resulting sequence of RL input states is numbered $(0, 1, 2, 3, 4)$.}
    \label{fig:token_based}
\end{figure}

Let $\text{VLM}_{\theta}$ represent the VLM architecture with all its parameters $\theta$ excluding the token output layer(s) at the end of the LLM. We will use $L_{\phi}$ to refer to the feed-forward language head that maps $\text{VLM}_{\theta}$'s outputs to a distribution over the tokens in our vocabulary $\mathcal{V}$. Our dataset contains turn-based transitions: $((o^{\text{text}}_t, o^{\text{img}}_t), a_t, r_{t+1}, d_{t+1}, (o^{\text{text}}_{t +1}, o^{\text{img}}_{t+1}))$, where $a_t$ is a complete reply and $d_{t+1}$ is one if we've reached the final turn of the interaction and zero otherwise. While it is simpler to view agent-environment interaction from a turn-based perspective, our training process will convert data to a token-based action space. The end result will be a more standard RL problem where states are the output of the VLM, and our policy decides between $|\mathcal{V}|$ discrete actions.
\begin{wrapfigure}{r}{.58\textwidth}
\vspace{-4mm}
\begin{center}
    \includegraphics[width=.57\textwidth]{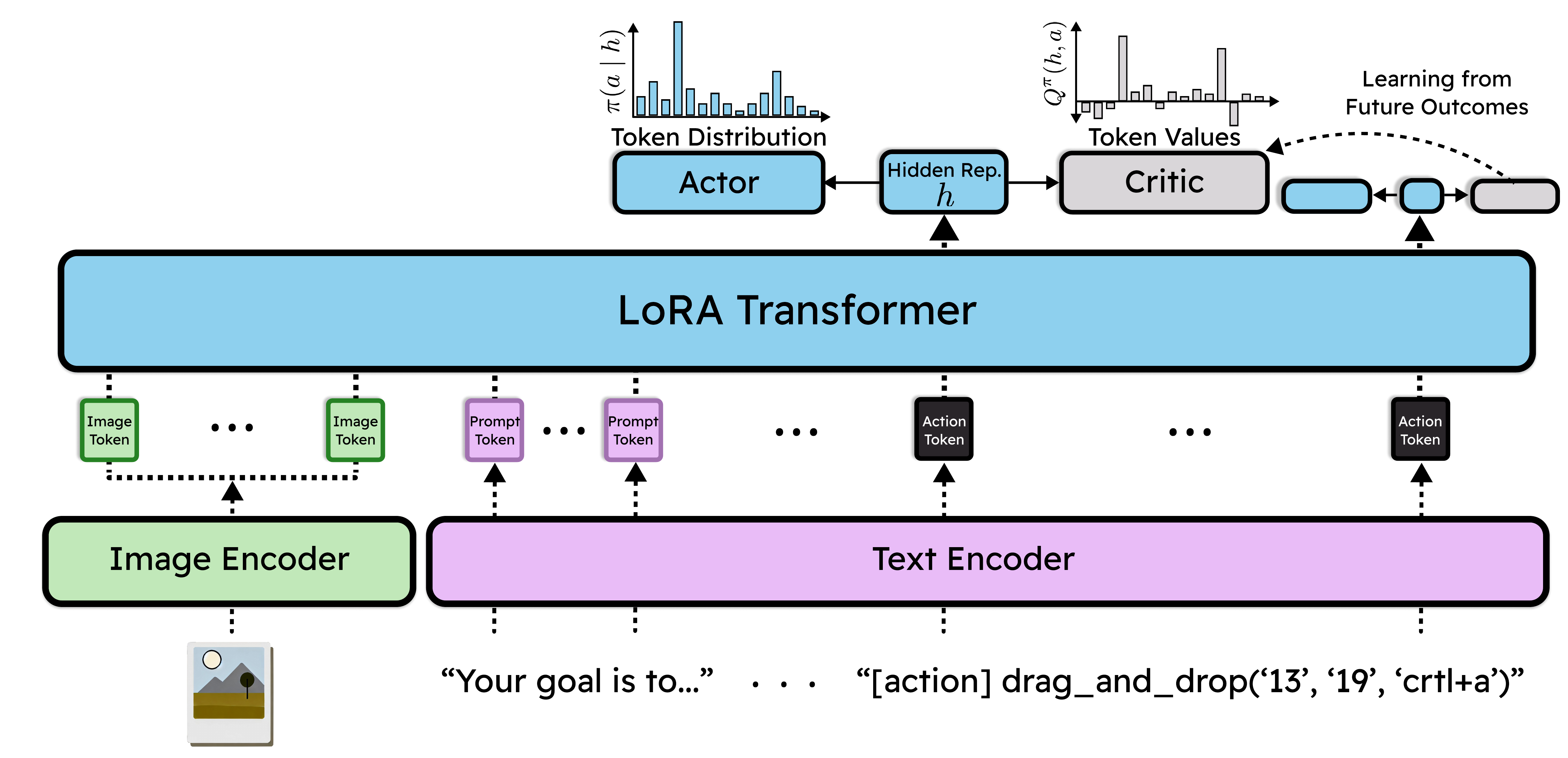}
    \caption{\textbf{VLM Actor-Critic.} We visualize a standard decoder-only VLM. Our method adds a second output head (critic) to estimate the future return achieved by selecting each token in the vocabulary. The critic filters the dataset of the language modeling head (actor).}
    \label{fig:method}
    \vspace{-5mm}
\end{center}
\end{wrapfigure}

\noindent\textbf{Converting from Turns to Tokens.} During inference, the VLM generated $a_t$ by tokenizing $(o^{\text{text}}_t, o^{\text{img}}_t)$ into a multi-modal sequence and auto-regressively continuing that sequence until a stop condition was met. Let $\tilde{a}$ be the tokenized action string with length $l$. The true input sequence to the VLM is $\tilde{a}$ concatenated to the end of a tokenized $(o^{\text{img}}, o^{\text{text}})$. $\text{VLM}_{\theta}$ outputs a sequence of $d$-dimensional vector representations where the last $l + 1$ indices correspond to the action decisions we want to optimize. We will summarize the process of retrieving the representations used to generate action tokens with the notation $\text{VLM}_{\theta}(o^{\text{img}}, o^{\text{text}}, a) = h \in \mathbb{R}^{(l + 1) \times d}$. Querying the VLM without the action text ($\text{VLM}_{\theta}(o^{\text{img}}, o^{\text{text}})$) returns the $d$-dimensional representation (of length one) that is used to begin the action reply. We can think of $\text{VLM}_{\theta}(o^{\text{img}_t}, o^{\text{text}}_t, a_t) = h_t$ as a sequence of RL states. The policy begins with $h_t^0$, selects a token action $\tilde{a}_t^0$, sees $h_t^1$, and receives no reward. At the end of the turn, the policy selects a token that terminates generation and returns control to the environment. From the policy's perspective the next observation is the first action query of the next turn, or $\text{VLM}_{\theta}(o^{\text{img}}_{t+1}, o^{\text{text}}_{t+1})$, with a reward and termination signal from the environment $r_{t+1}$ and $d_{t+1}$. Following this logic, we can convert a turn-based transition $((o^{\text{text}}_t, o^{\text{img}}_t), a_t, r_{t+1}, d_{t+1}, (o^{\text{text}}_{t +1}, o^{\text{img}}_{t+1}))$ to token-based RL transitions by computing:

\vspace{-3mm}

\begin{minipage}{0.45\textwidth}
  \begin{align}
    h &= \text{VLM}_{\theta}(o^{\text{img}}_t, o^{\text{text}}_t, a_t) \label{eq:state_pass}
    \end{align}
\end{minipage}
\begin{minipage}{0.45\textwidth}
    \begin{align}
      h' &= \cancel{\nabla} \text{VLM}_{\theta}(o^{\text{img}}_{t+1}, o^{\text{text}}_{t+1}) \label{eq:next_state_pass}
    \end{align}
\end{minipage}

Where $\cancel{\nabla}$ is a stop-gradient that puts the VLM in inference mode. The expanded batch of RL transitions is: $(h^0, \tilde{a}_t^0, r^1 = 0, d^1 = 0, h^1), (h^1, \tilde{a}_t^1, r^2 = 0, d^2 = 0, h^2), \dots, (h^{l}, \tilde{a}_t^{l}, r^{l + 1} = r_{t + 1} , d^{l + 1} = d_{t+1}, h^{l + 1} = h')$. This process is visualized in Figure \ref{fig:token_based}. After the conversion, we can apply RL exactly as we would in any other setting, and gradients will flow back to the base VLM.

\noindent\textbf{Language Loss.} SFT is the most popular approach to VLM agent fine-tuning. In multi-turn SFT, we compute the next-token prediction error causally along the entire sequence and mask the loss so that we are only maximizing the likelihood of the agent's replies. In our notation, the SFT loss for turn $i$ would be written:

\vspace{-9mm}

\begin{align}
    \mathcal{L}_{\text{SFT}}(i) &= -\frac{1}{l} \sum_{j=0}^{l} \log L_{\phi}(\tilde{a}_i^j \mid h_i^j) 
    \label{eq:sft}
\end{align}

\vspace{-2mm}

RL would call Equation \ref{eq:sft} behavior cloning, and its limitation is that it imitates every decision in the dataset and, therefore, cannot improve beyond its performance. We need to \textit{filter} the dataset with some function $f(h, a) \in \{0, 1\}$ to avoid imitating sub-optimal decisions:

\vspace{-5mm}

\begin{align}
    \mathcal{L}_{\text{FSFT}}(i) &= -\frac{1}{l}\sum_{j=0}^{l} f(h_i^j, \tilde{a}_i^j)\log L_{\phi}(\tilde{a}_i^j \mid h_i^j) 
    \label{eq:fsft}
\end{align}

\vspace{-3mm}

The filter creates a second mask for action tokens that should not be imitated. While we could create a heuristic filter by masking low-return trajectories, we can let RL outperform heuristics by learning a filter from data. We create a second output head, $Q_{\psi}$, which takes the VLM representations as input and produces $|\mathcal{V}|$ outputs --- much like the language head $L_{\phi}$. Let $Q_{\psi}(h)[k]$ be the output corresponding to token $k$. Assuming $Q_{\psi}$ outputs correct $Q$-values (Section \ref{sec:background}), the filter can become:

\vspace{-4mm}

\begin{align}
    f(h_i^j, \tilde{a}_i^j) = Q_{\psi}(h_i^j)[ \tilde{a}_i^j] > Q_{\psi}(h_i^j)^\intercal L_{\phi}(h_i^j) + \beta
    \label{eq:filter}
\end{align}

\vspace{-1mm}

In RL terminology, the filter approves actions that have an \textit{advantage} of at least $\beta$: $f(s, a) = Q^{\pi}(s, a) - V^{\pi}(s) > \beta$. We will refer to Equation \ref{eq:fsft} with the filter in Equation \ref{eq:filter} as Advantage Filtered SFT (AFSFT).

\noindent\textbf{Value Loss.} Our filter will only be useful if we can train $Q_{\psi}$ to output accurate $Q$-values. We train $Q_{\psi}$ with a one-step temporal difference (TD) loss where we bootstrap an estimate of $Q(s, a)$: 
$y^i = r^{i+1} + \gamma (1 - d^{i+1}) \cancel{\nabla}(Q_{\psi}(h^{i+1})^\intercal L_{\phi}(h^{i+1}))$. For stability, $y$ uses a moving average of the critic parameters $\psi$ \citep{lillicrap2015continuous}. We update our current predictions based on the improved estimate:

\vspace{-6mm}

\begin{align}
    \mathcal{L}_{\text{TD}}(i) &= \frac{1}{l}\sum_{j=0}^{l}(Q_{\psi}(h_i^j)[\tilde{a}_i^j] - y_i^{j})^2
    \label{eq:critic}
\end{align}

\vspace{-2mm}

\noindent\textbf{Training.} We can now put these RL components together to update our VLM. Figure \ref{fig:method} provides a high-level overview. Each training step samples a batch of turn-based transitions from our dataset $\mathcal{D}$. We convert VLM agent data to a sequential RL problem with one forward pass of the VLM on turn $i$ (Equation \ref{eq:state_pass}) and one inference-only pass on the following turn $i+1$ (Equation \ref{eq:next_state_pass}). We then compute the critic loss (Equation \ref{eq:critic}) and actor loss (Equations \ref{eq:fsft} and \ref{eq:filter}), and take a gradient step on the joint loss: 

\vspace{-6mm}

\begin{align}
    \mathcal{L}_{\text{VLMQ}} = \displaystyle \mathop{\mathbb{E}}_{i \sim \mathcal{D}} \left[ \mathcal{L}_{\text{FSFT}}(i) + \lambda \mathcal{L}_{\text{TD}}(i) \right]
    \label{eq:vlmq}
\end{align}
\vspace{-3mm}

Note that the use of stop-gradients in TD targets $y$ and the filter $f$ mean that the actor (language head) $L_{\phi}$ only optimizes $\mathcal{L}_{\text{FSFT}}$ while the critic $Q_{\psi}$ only optimizes $\mathcal{L}_{\text{TD}}$. However, the VLM trains on both objectives simultaneously and balances their values with a hyperparameter $\lambda$. We use standard techniques to save GPU memory during optimization, including training in mixed precision with an $8$-bit optimizer \citep{dettmers20218}. We also use LoRA to adapt the VLM's behavior with a small number of trainable parameters \citep{hulora}. However --- unlike the other efficiency techniques --- LoRA plays an important conceptual role and would likely still be useful even if hardware resources were not a concern. Optimizing the VLM on task-specific data with $\mathcal{L}_{\text{VLMQ}}$ will almost surely degrade its performance on other tasks. In theory, $\mathcal{L}_{\text{FSFT}}$ lets us retain general behavior by imitating reward-free SFT tokens from a standard instruction-tuning dataset alongside RL training, but this would be expensive in practice. LoRA makes this unnecessary by letting us discard the weight adapters and recover the base model at any time during deployment. 

\noindent\textbf{Extending SFT.} Our base VLM optimizes a policy loss that is a safe improvement over SFT in most cases and lets us recover SFT if necessary. When the learned filter $f$ is noisy or near its initialization, Eq. \ref{eq:fsft} effectively becomes SFT on random subsets of the dataset. When $Q$-value estimation is successful, Eq. \ref{eq:fsft} creates a policy that outperforms its training data by declining to imitate action token outputs thought to be sub-optimal. There are some concerns when training on expert datasets where the dataset is of uniformly high quality, as the margin of advantages used to compute Eq. \ref{eq:fsft} would be close to zero. If we know that our dataset contains optimal decisions, we could manually recover standard SFT with $\beta = -\infty$ and $\lambda = 0$. Like SFT, but unlike many foundation model RL actor-critics, $\mathcal{L}_{\text{VLMQ}}$ requires a single base VLM.

\noindent\textbf{Offline-to-Online Fine-Tuning.} In applications like robotics, offline pre-training is often motivated by safety or cost reduction. Many VLM agent applications involve safe and inexpensive simulated environments, but learning from existing datasets will be important for two different reasons. First, it will help us overcome the challenge of outputting correct action syntax without a separate SFT stage. We can use more capable models or other techniques to generate a diverse dataset of valid action syntax but do not need to assume those actions are high-quality semantic decisions. The token-based action format allows agents to copy syntax details from poor actions without imitating the entire semantic decision. The second benefit addresses token vocabulary ($|\mathcal{V}|$) action spaces that are orders of magnitude larger than we would typically see in discrete actor-critics (Table \ref{tbl:vlm_intro}). Many token actions are not relevant to the task and will never be selected or appear in our dataset \citep{snelloffline}. The outputs of $Q_{\psi}$ for these tokens will never be optimized and should effectively be treated as random numbers that do not represent meaningful value estimates. Many offline RL techniques manage this miscalibration by constraining updates to actions that appear in their (small) static dataset. Token action spaces are so large that they create a unique situation where this constraint is necessary even when we do allow for online interaction. 

%$\mathcal{L}_{\text{FSFT}}$ is compatible with both offline and online RL \cite{wang2020critic, ferret2021self}. This makes it well-suited to applications where we use an offline dataset to learn a reasonable initial policy before interacting with the environment \cite{nair2018overcoming, nair2020awac}. 
\vspace{-2mm}
\section{Experiments}

\vspace{-5mm}

\begin{figure}[h!]
    \centering
    \includegraphics[width=.75\linewidth]{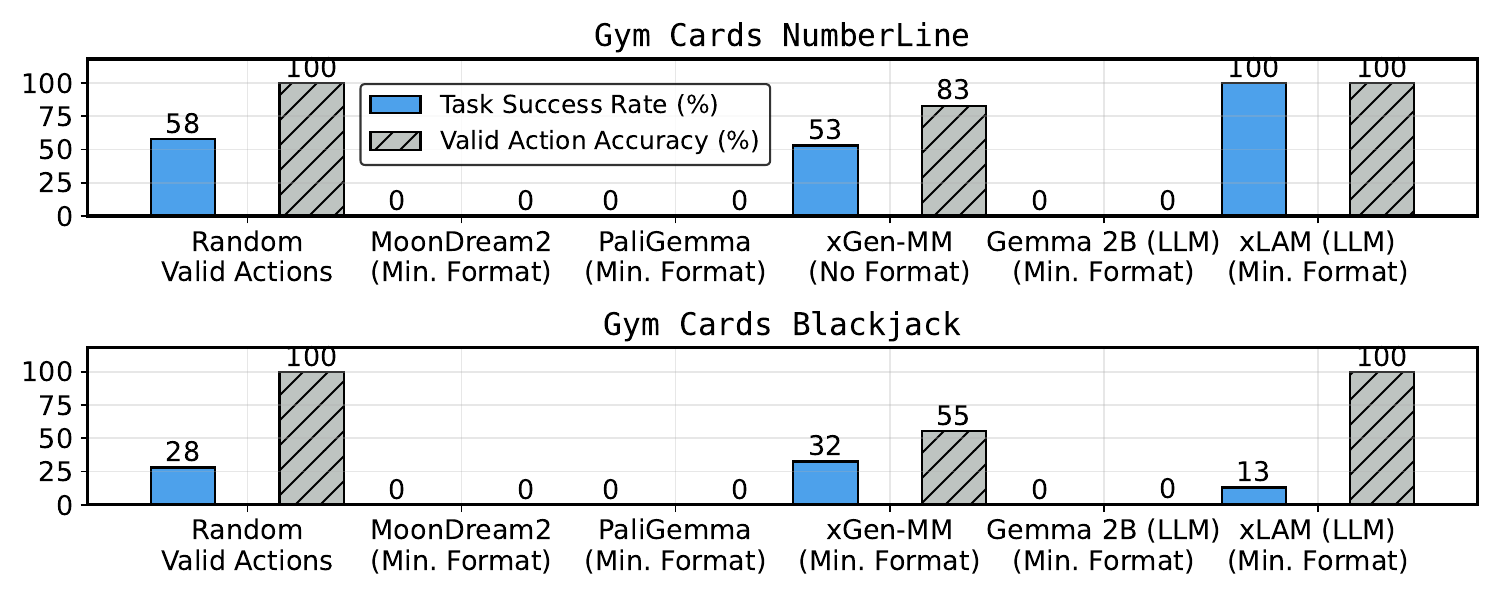}
    \vspace{-5mm}
    \caption{\textbf{Gym Cards Model Evaluations.} We compare the task success and the rate of valid model output syntax of base models in two Gym Cards environments. LLM prompts replace the image with an equivalent text description.}
    \label{fig:gym_cards_base_models}
\end{figure}

We begin by highlighting the difficulty of prompting base VLMs to succeed in simple agent-like tasks. Gym Cards \citep{zhai2024fine} converts toy RL decision-making tasks like the game of Blackjack into testbeds for VLM-Agent research. The true action space $\mathcal{A'}$ is a small set of discrete choices, but following \citep{liu2024visual}, we make this domain more representative of the broader problems involved in prompting the model to output specific syntax by setting $\texttt{parse}_{\text{env}}$ to match a text action with only light corrections. We choose the $\texttt{parse}_{\text{agent}}$ syntax of ``\texttt{[action]}'' because it is trivial to prompt the base model to pass this check by including the action string at the beginning of the reply. Additional training and action parsing details are provided in Appendix \ref{app:implementation}. We follow a general prompting approach summarized in Section \ref{sec:prompting}, where a text input includes instructions about the objective of the task and the expected syntax of the models' action replies (with an example). Appendix \ref{app:prompting} contains example prompts for our experiments.

\begin{figure}
    \centering
    \includegraphics[width=\textwidth]{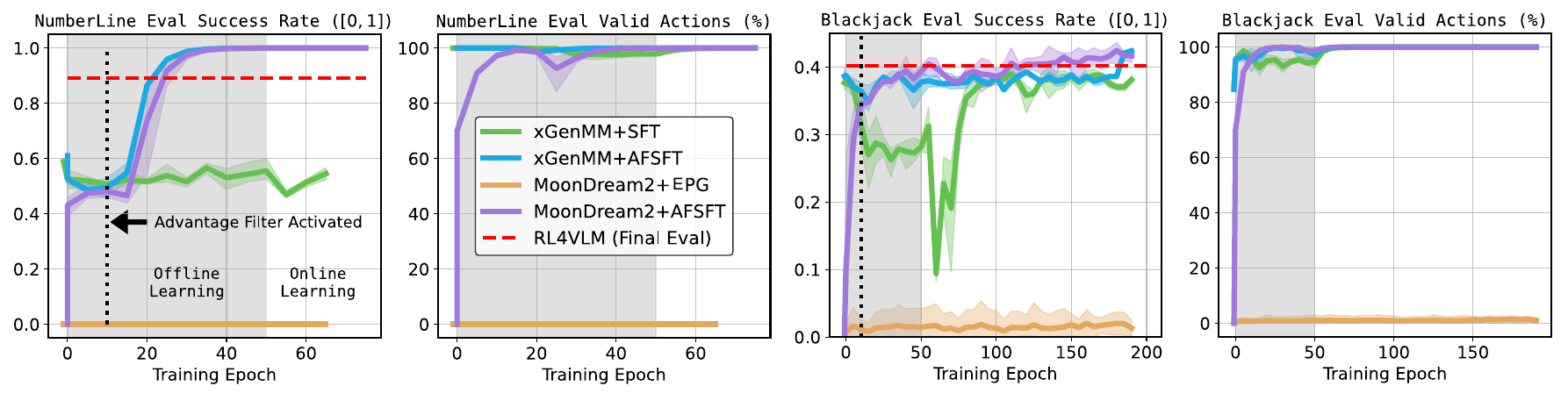}
    \vspace{-8mm}
    \caption{\textbf{Offline-to-Online Gym Cards.} \textbf{(Left, Center Right)} Task success rate ($[0, 1]$) compared against a reference score for a similar setup in \citet{zhai2024fine}. A \textcolor{gray}{gray} background indicates offline learning on a fixed dataset. A white background indicates that online environment interaction is enabled. \textbf{(Center Left, Right)} Action syntax accuracy during evaluation.}
    \label{fig:gym_cards_offline_to_online}
    \vspace{-7mm}
\end{figure}

Figure \ref{fig:gym_cards_base_models} compares the performance of the three base VLMs in Table \ref{tbl:vlm_intro} using only prompting and without further fine-tuning. We see that it is surprisingly difficult to get these smaller open-weight VLMs to cooperate with the expected action syntax. Action syntax is closely related to tool-use (Section \ref{sec:background}), where larger models tend to be more accurate, and this capability is more reliable in larger LLMs. We can demonstrate this by using privileged environment data to add a text summary of the image to the observation prompt. LLMs can then be used in a nearly equivalent setup by simply ignoring the visual input. xLAM \citep{zhang2024xlam} --- an LLM specifically targeted at tool-use and function-calling applications --- achieves high accuracy. Off-policy RL fine-tuning will let us learn from the interactions of larger models (including LLMs with alternate text prompts). However, even when models have reasonable syntax accuracy, they might only output certain subsets of the action space or make incorrect semantic decisions that will not solve the task. Appendix \ref{app:implementation} Figure \ref{fig:additional_prompting} repeats this experiment in the BabyAI \citep{chevalier-boisvert2018babyai} domain and explores the addition of ``thought''-style prompting. We find all VLMs output inaccurate syntax, and only xGen-MM has a non-zero success rate.

\begin{wrapfigure}{r}{.67\textwidth}
\begin{center}
    \centering
    \vspace{-10mm}
    \includegraphics[width=.69\textwidth]{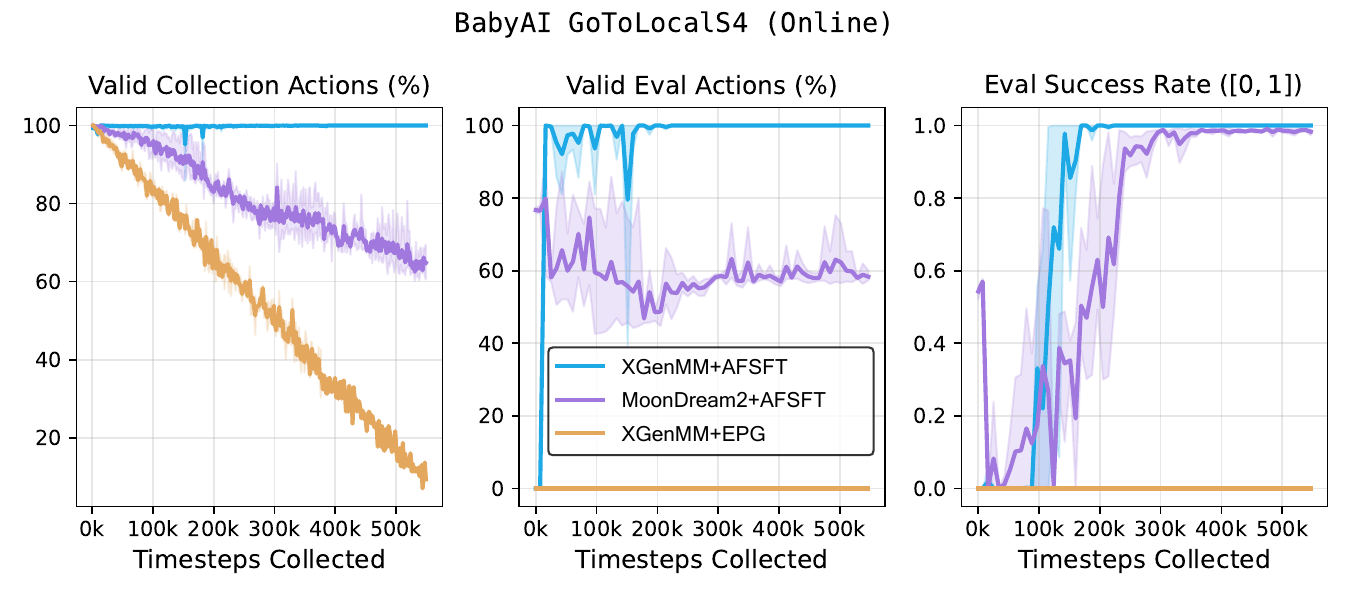}
    \vspace{-8mm}
    \caption{\textbf{Online RL in BabyAI.} An $\epsilon$-greedy exploration schedule allows for online learning by introducing valid action syntax that can be refined by the critic filter to produce a successful policy.}
    \label{fig:babyai}
    \vspace{-3mm}
\end{center}
\end{wrapfigure}
We can improve performance by taking these same VLM-agents and fine-tuning them with $\mathcal{L}_{\text{VLMQ}}$ (Eq. \ref{eq:vlmq}). Figure \ref{fig:gym_cards_offline_to_online} demonstrates the technique in two Gym Cards tasks. We initialize a dataset $\mathcal{D}$ to the actions generated by a random policy; this data makes sub-optimal (random) decisions that are properly formatted as text. We also mix in data collected by prompting base VLMs --- which are both inaccurate and low-performance. More detail about the initial offline datasets used in our experiments is provided in Appendix \ref{app:offline_datasets}. For demonstration purposes, the advantage threshold is initialized to $\beta = -\infty$ (Eq. \ref{eq:filter}) such that the language head (RL actor) is optimizing SFT on this sub-optimal dataset while beginning to optimize the critic value regression loss (Eq. \ref{eq:critic}). This quickly brings both xGenMM and MoonDream2 to a similar performance level. After a few hundred gradient steps, we schedule the advantage threshold to reach $\beta = 0$; the critic is now filtering sub-optimal tokens within each turn's reply, and the agent's performance improves to match or exceed the RL4VLM \citep{zhai2024fine} reference score. Because $\mathcal{L}_{\text{VLMQ}}$ is equally applicable online and offline, we are free to use our current RL agent to collect additional data. In this case, the agents have already converged to high performance but maintain this level as the buffer fills with fresh data. In RL, it is not uncommon to see a dip in performance during the transition between offline and online learning due to distribution shift and related optimization challenges \citep{zhang2024perspective}, but Eq. \ref{eq:fsft} avoids many of these obstacles, and we do not observe instability.

Figure \ref{fig:gym_cards_offline_to_online} includes results from an alternative actor update that replaces Eq. \ref{eq:fsft} with a more standard online objective that directly maximizes $Q$-values according to the current critic: $\mathcal{L}_{\text{EPG}}(i) =  -\frac{1}{l} \sum_{j=0}^{l} L_{\phi}(h_i^j)^{\intercal} \cancel{\nabla} Q_{\psi}(h_i^j)$ \citep{JMLR:v21:18-012, christodoulou2019soft}. While this objective is effective in discrete actor-critics in more typical RL settings, it is not effective here due to 
miscalibrated $Q$-values in the critic network's output space resulting from large token vocabularies.

\begin{figure*}[t!]
    \centering
    \vspace{-3mm}
    \includegraphics[width=0.95\linewidth]{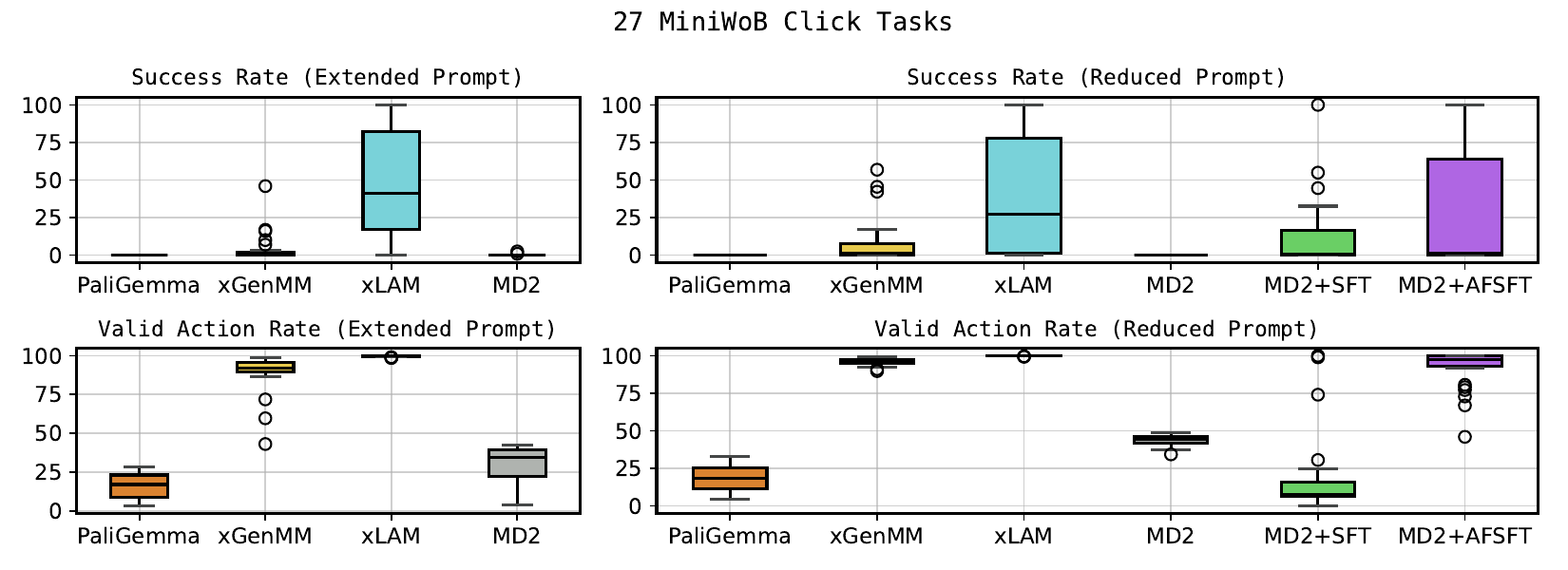}
    \vspace{-2mm}
    \caption{\textbf{Offline RL in BrowserGym MiniWoB.} Box plots indicate the median and interquartile range across $27$ ``click'' tasks. Models are evaluated with and without an ``extended prompt'' that includes the webpage DOM and language descriptions of the code-like action space (Appendix \ref{app:prompting}). Individual task results are listed in Tables \ref{tbl:miniwob_suc} and \ref{tbl:miniwob_acc}. ``MD2'' abbreviates MoonDream2.}
    \label{fig:miniwob_main_text}
    \vspace{-6mm}
\end{figure*}

Training an RL policy from a fully offline dataset more closely resembles the setup of SFT agent fine-tuning, where the ability to collect online data to improve performance is a convenient feature. However, we could also fine-tune the VLM entirely online using a buffer of its own experience. Figure \ref{fig:babyai} demonstrates online learning in a BabyAI task. BabyAI is another domain where the true action space is a small set of discrete choices, which allows for a simple exploration strategy where we can replace the model's response with a random (valid) decision with probability $\epsilon$ that decays over time. Because the dataset then contains some valid actions that avoid the syntax reward penalty, the AFSFT update learns to favor them over the model's own (inaccurate) outputs, leading to a spike in action accuracy. From there, the $ Q$ filter can prefer actions that lead to task success over other valid but sub-optimal decisions, and the VLM's success rate rises to nearly $100\%$.

In domains where the true action space is open-ended text or state-dependent code, it is more difficult to generate valid exploration actions. We evaluate this with the BrowserGym \citep{drouin2024workarena} variant of the MiniWoB++ benchmark \citep{pmlr-v70-shi17a, liu2018reinforcement}. MiniWoB is a suite of locally hosted browser tasks for RL agents. BrowserGym creates a standardized interface for web tasks with code-like action syntax (Figure \ref{fig:miniwob_prompt_pt1}). We use $3$ VLMs and the xLAM LLM to collect a dataset of $340$k actions across $27$ different ``click'' navigation tasks. Figure \ref{fig:miniwob_main_text} summarizes these models' action accuracy and task success rate. We compare the performance of a more detailed prompt that includes HTML text in addition to a browser screenshot against a simplified variant that leads to more affordable input lengths. Our dataset is partially generated by filling the agent's action reply with the beginning of a randomly selected BrowserGym function call --- adding many inaccurate or irrelevant browser commands to the dataset and creating a challenging offline RL problem. We then fine-tune the worst performing VLM (MoonDream2) using both SFT (Eq. \ref{eq:sft}) and AFSFT (Eq. \ref{eq:vlmq}). Figure \ref{fig:miniwob_main_text} highlights $\mathcal{L}_{\text{VLMQ}}$'s ability to sift through noisy datasets in order to recover a MoonDream2 policy that more closely resembles the larger xLAM model.
\section{Conclusion}

The standard approach to training VLM agents typically involves using a larger, more capable model to generate a dataset, which is then used to train a smaller model via supervised fine-tuning (SFT). In this work, we show that advantage-filtered supervised fine-tuning (AFSFT) --- an offline reinforcement learning technique that masks actions predicted to reduce downstream performance --- can effectively replace SFT in scenarios where datasets are suboptimal or action formats are unreliable. AFSFT enables a smooth transition between learning from static offline data and interacting with the environment online, while requiring only a minimal architectural change: the addition of a secondary token output head atop a fine-tunable base model.

\bibliography{iclr2025_conference}
\bibliographystyle{iclr2025_conference}

\appendix
\section{Appendix}

\subsection{Additional Related Work}
\label{app:additional_related}

RL is used in the post-training process to realign a base model with human feedback (RLHF) and enforce safety measures \citep{NEURIPS2022_b1efde53, kaufmann2023survey, bai2022training}. RLHF typically uses online (on-policy) updates based on constrained policy gradient techniques \citep{zhu2023fine, ahmadian2024back} that prevent the fine-tuned policy from over-optimizing a learned reward model of human preferences \citep{lambert2024rewardbench}. Unlike the more general ``agent'' setting, RLHF interactions typically last for one turn --- a simplification that allows for alternate updates like DPO \citep{rafailov2024direct}. VLMs can also be used to create reward functions for training smaller RL policies from scratch \citep{rocamonde2024visionlanguage, wang2024rl}. Our work updates the VLM to output decisions directly, which is useful for text-based actions because learning to output language from a randomly initialized policy via RL is sample inefficient. Another category of methods fine-tunes foundation models as policies while avoiding the action syntax challenge that comes with text actions (Section \ref{sec:background}). When the environment has a small discrete set of valid actions, we can iterate over our options and compare their likelihood according to the LLM \citep{carta2023grounding}. We could also discard text outputs and use the hidden state of the base model to train a new action output layer from scratch \citep{chen2024vision, szot2023large, xu2024language}.

\subsection{Additional Implementation Details}
\label{app:implementation}

\subsubsection{Training}

Table \ref{tbl:models_appendix} lists key hyperparameters and version control details for fine-tuning the MoonDream2 and xGen-MM models (with SFT or RL). Our LoRA \citep{hulora} settings are based on recommendations from SFT fine-tuning pipelines by the models' authors.

\begin{table}[h!]
\centering
\resizebox{.7\linewidth}{!}{
\begin{tabular}{@{}lcc@{}}
\toprule
                                  & \textbf{MoonDream2} & \textbf{xGen-MM}                                                 \\ \midrule
Version Notes                     & revision 2024-05-20 & xgen-mm-phi-3-mini-instruct-r-v1                                 \\
Precision                         & bf16                & bf16                                                             \\ \midrule
\multicolumn{1}{c}{\textit{LoRA}} &                     &                                                                  \\ \midrule
r                                 & 8                   & 8                                                                \\
$\alpha$                          & 16                  & 16                                                               \\
Dropout                           & .05                 & .05                                                              \\
Param Groups                      & ``mixer.Wqkv"       & {[}``qkv\_proj", ``q\_proj", ``gate\_up\_proj", ``down\_proj"{]} \\
Tunable Params                    & 1,572,864           & 12,582,912                                                       \\ \bottomrule
\end{tabular}}
\caption{\textbf{VLM Model Training Details.}}
\label{tbl:models_appendix}
\end{table}

Table \ref{tbl:hparams} provides a full listing of the hyperparameters involved in collecting environment data and optimizing the VLM on the actor-critic RL loss (Eq. \ref{eq:vlmq}).

\begin{table}[h!]
\centering
\resizebox{.95\linewidth}{!}{
\begin{tabular}{@{}lccc@{}}
\toprule
                                                                                                                           & \textbf{Gym Cards}                 & \textbf{BabyAI}                    & \textbf{MiniWoB}    \\ \midrule
\multicolumn{1}{c}{\textit{Exploration / Language Sampling}}                                                               &                                    &                                    &                     \\ \midrule
Language Temperature During Exploration                                                                                    & .4                                 & .4                                 & .5                  \\
Max Reply Tokens                                                                                                           & 24                                 & 24                                 & 32                  \\
\begin{tabular}[c]{@{}l@{}}Expl. $\epsilon$ Schedule (Start $\rightarrow$ Stop), \\ {[}Over Online Epochs{]}\end{tabular} & (.35 $\rightarrow$ .05), {[}50{]} & (1.0 $\rightarrow$ .1), {[}300{]} & N/A (Offline)       \\
Explore Action Prompt                                                                                                      & random                             & random                             & N/A                 \\
Exploit / Eval Action Prompt                                                                                               & agent                              & agent                              & agent               \\ \midrule
\multicolumn{1}{c}{\textit{Optimization}}                                                                                  &                                    &                                    &                     \\ \midrule
Minibatch Size Per GPU (A100 40GB)                                                                                         & 3                                  & 3                                  & 1                   \\
Parallel GPUs                                                                                                              & 2                                  & 2                                  & 1                   \\
Gradient Accumulation                                                                                                      & 4                                  & 4                                  & 8                   \\
Learning Rate (VLM)                                                                                                        & 1e-4                               & 1e-4                               & 1e-4                \\
Learning Rate (Critic MLP)                                                                                                 & 6e-4                               & 8e-4                               & 7e-4                \\
Grad Norm Clip                                                                                                             & 2                                  & 2                                  & 2                   \\
VLM Linear LR Warmup Steps                                                                                                 & 250                                & 500                                & 300                 \\
Critic Linear LR Warmup Steps                                                                                              & 100                                & 50                                 & 100                 \\ \midrule
\multicolumn{1}{c}{\textit{Learning Schedule}}                                                                             &                                    &                                    &                     \\ \midrule
Buffer Min Size Before Training (Per GPU)                                                                                  & N/A (Offline Demos)                & 5k                                 & N/A (Offline Demos) \\
Max Buffer Size (Per GPU)                                                                                                  & 50k                                & 100k                               & N/A (Offline)       \\
Parallel Environments (Per GPU)                                                                                            & 6                                  & 6                                  & 8                   \\
Start Collecting on Epoch                                                                                                  & 50                                 & 0                                  & $\infty$            \\
Timesteps Collected Per Env Per Epoch                                                                                      & 96                                 & 150                                & 0                   \\
Grad Updates Per Epoch                                                                                                     & 512                                & 512                                & 3072                \\
Total Epochs                                                                                                               & 75 (NumberLine) 200 (Blackjack)    & 600                                & 50                  \\ \midrule
\multicolumn{1}{c}{\textit{TD Learning}}                                                                                   &                                    &                                    &                     \\ \midrule
Target Critic $\tau$                                                                                                       & .008                               & .008                               & .009                \\
Critic Loss Weight $\lambda$                                                                                               & 1.0                                & 8.0                                & 1                   \\
Critic MLP Hid. Dim. & $d_{\text{model}}$ & $d_{\text{model}}$ & 400 \\
$\gamma$ & .995 & .995 & .995 \\
PopArt $\beta$                                                                                                             & 5e-4                               & 5e-4                               & 5e-4                \\
PopArt Init $\nu$                                                                                                          & 100                                & 100                                & 100                 \\ \bottomrule
\end{tabular}}
\caption{\textbf{Training Hyperparameters.} PopArt \citep{vanhasselt2016learningvaluesordersmagnitude} is an RL implementation detail that helps normalize $Q$-value predictions and reduces tuning of the critic loss weight ($\lambda$) and learning rate.}
\label{tbl:hparams}
\end{table}

\subsubsection{Action Prompting and Parsing}
\label{app:action_prompting}

\textbf{Action Parsing.} The goal of our method is to be a flexible replacement for SFT updates in an agent setting. Therefore, we need our RL update to be compatible with arbitrary prompting schemes, especially chain-of-thought \citep{wei2022chain} or ReACT-style \citep{yao2022react} reasoning. We can allow for this by assuming a method $\texttt{parse}_{\text{agent}}$ (Fig. \ref{fig:fig1}) can modify the model's text output before it is passed to the environment. In this work, we take a simple approach and remove all text up to and including the special tag ``\texttt{[action]}''. This allows the model room to reason in open-ended text before selecting its action. We use ``\texttt{[action]}'' instead of more common json agent formats because we are evaluating small-scale models that struggle to output the requested syntax, and we can easily prompt the agent with an ``\texttt{[action]}'' tag.

In order for the environment simulator to execute the agent's actions, it needs to be able to parse the intended action from open-ended language outputs ($\texttt{parse}_{\text{env}}$ in Figure \ref{fig:fig1}). In domains like the BrowserGym version of MiniWoB (and many real-world applications of closed-source LLM/VLMs), actions are interpreted as function calls with state-dependent arguments. Even though the underlying decision-making problem of Gym Cards and BabyAI can be represented by discrete actions, we follow the lead of \citet{zhai2024fine} in simulating an action syntax barrier by requiring the model to output a string from a set of discrete choices in the prompt (Figure \ref{fig:gym_cards_prompt}). In Gym Cards, we require an exact match after removing any whitespace and markdown tags. In BabyAI, we are slightly more lenient and parse the first of any match in the output string.

When the model produces syntax outside of the expected format, it cannot advance the simulator and collect useful experience, even if it subjectively demonstrates some understanding of the task. This dilemma is common in LLM evaluations and can sometimes be the root cause of ``emergent'' behavior \citep{wei2022emergentabilitieslargelanguage}: models that are only a slight improvement in terms of training loss or human preferences can offer a dramatic improvement in automated evaluations because they cross the somewhat arbitrary threshold of being able to interpret the requested format. As discussed in Section \ref{sec:background}, this capability has been emphasized and improved by tool-use benchmarks. However, it is an especially difficult problem in RL because invalid action syntax prevents us from reaching environment rewards and creates a sparse-reward exploration problem. The space of possible text actions is so large that default RL strategies, like randomly sampling tokens from the action space, are unlikely to find valid behavior. At the same time, the base model may be biased towards certain inaccuracies, such that sampling with higher temperature is unlikely to discover the successful examples necessary for RL to correct its mistakes.

\textbf{Action Prompting.} Action prompting refers to strategies that modify the \texttt{prompt} method (Figure \ref{fig:fig1}) to induce more accurate action syntax or a larger variety of decisions (or both). In the main text and appendix, ``No Format'' or ``agent'' formatting refers to the natural default of letting the model respond to the observation solely by sampling from the outputs of its own language head (policy). ``Minimum Format'' tries to overcome the most obvious bottleneck of not outputting the ``\texttt{[action]}'' tag by beginning the agent's reply with a tokenized version of that string. ``Random Valid Actions`` or ``random prompting'' selects random actions from the environment's underlying action space $\mathcal{A}'$ (Section \ref{sec:vlm_rl}) and maps them to valid text. In BrowserGym, this strategy is not practical because actions require knowledge of HTML element IDs and open-ended text. Instead, we use a ``partial'' prompting strategy where we sample from prefixes of valid actions. More specifically, we can sample method names from the browser interface and have the agent attempt to complete the arguments for that call. These prompting strategies create a version of $\epsilon$-greedy exploration in RL that is used during online learning and dataset collection. With probability $\epsilon$, we can sample actions with an exploration strategy (e.g., a random valid action). Otherwise, we can use an exploitation strategy (e.g., the agent's greedy output).

\begin{figure}[h!]
    \centering
    \includegraphics[width=\linewidth]{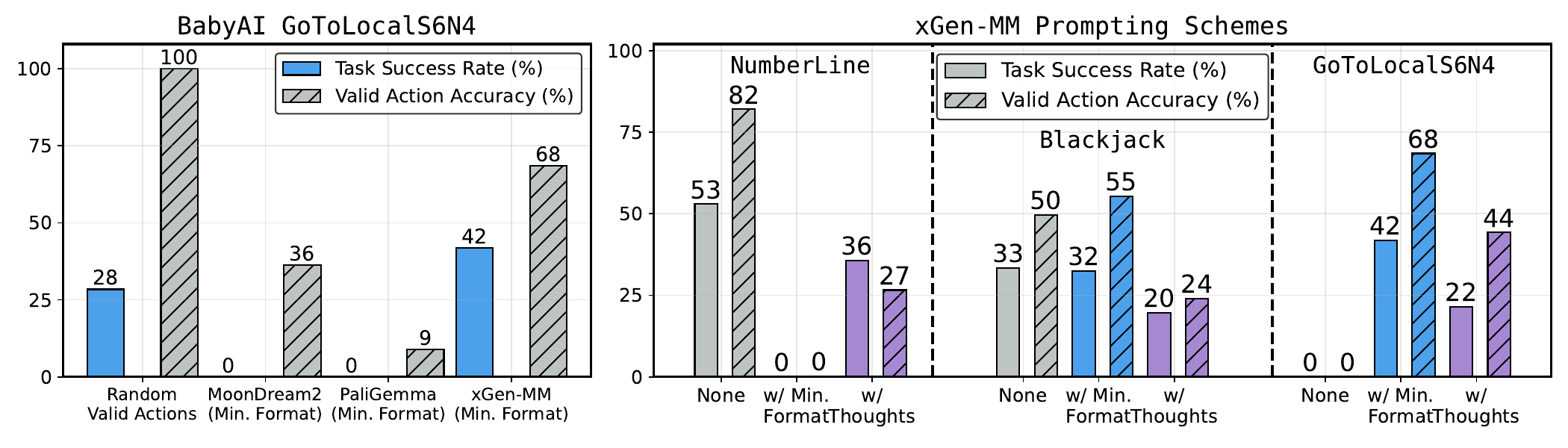}
    \vspace{-5mm}
    \caption{\textbf{Base Model Evaluations and Prompting.} (Left) We repeat the base model action syntax and success rate evaluation in a BabyAI domain. (Right) Comparing basic prompt variants with xGenM-MM. ``w/ Min. Format'' adds assistance with action parsing syntax. ``w/ Thoughts'' adds ReACT-style \cite{yao2022react} reasoning.}
    \label{fig:additional_prompting}
\end{figure}

\subsection{Environment Details}

\subsubsection{Reward Functions}

Our problem formulation assumes multi-turn interactions where rewards are given after every action decision. We can take advantage of this by adding reward terms beyond the reward function of the inner environment --- which is often a binary indicator of success or failure at the end of the episode. We add per-timestep penalties for outputting invalid action choices in order to encourage more useful data collection (Appendix \ref{app:action_prompting}). We also introduce time limits and penalties for running out of time in the environment or getting stuck in a cycle of several invalid actions. Modifications to the standard reward functions of each environment are detailed in Table \ref{tbl:rewards}.

\label{app:rewards}

\begin{table}[h!]
\centering
\resizebox{.65\linewidth}{!}{
\begin{tabular}{l|c|c|c|}
\cline{2-4}
\multicolumn{1}{c|}{}                                        & \textbf{Gym Cards} & \textbf{BabyAI} & \textbf{MiniWoB} \\ \hline
\multicolumn{1}{|l|}{Inner Env Reward Multiplier}                        & $\times 1$                 & $\times 100$            & $\times 1$               \\ \hline
\multicolumn{1}{|l|}{On Invalid Action}                      & -2                 & -1              & -2               \\ \hline
\multicolumn{1}{|l|}{Timeout on Consecutive Invalid Actions} & 3                  & 5               & 5                \\ \hline
\multicolumn{1}{|l|}{Timeout on Total Steps}                 & 10                 & 64              & 25               \\ \hline
\multicolumn{1}{|l|}{On Timeout}                             & -10                & -200            & -10              \\ \hline
\end{tabular}}
\caption{\textbf{Reward Terms.} Reward functions are combinations of their environment's natural reward signal (in this case: a success/failure indicator given at the end of an episode) and per-timestep signals to encourage correct action formatting.}
\label{tbl:rewards}
\end{table}

\subsubsection{Offline Datasets}

\begin{itemize}
    \item \textbf{Gym Cards.} Each of the Blackjack and NumberLine initial (offline) datasets contains $28800$ transitions. $24$k transitions were generated by random valid actions while $1.6$k were the outputs of MoonDream2 without any action action formatting. Both domains collected $2.4$k actions from xGen-MM. Blackjack assisted xGen-MM with minimum action formatting, while in NumberLine, we could reach higher performance without action formatting (Figure \ref{fig:gym_cards_base_models}) due to a tokenization quirk. VLM sampling used a temperature of $0.4$.
    \item \textbf{MiniWoB.} The MiniWoB dataset contains $364320$ transitions spread approximately evenly across $27$ tasks (Table \ref{tbl:miniwob_suc}). With probability $\epsilon = .5$, a partial action was generated to encourage a diverse dataset (Appendix \ref{app:action_prompting}). Otherwise, actions were sampled from the base VLM/LLM with temperature $0.5$ (without formatting assistance). PaliGemma, xGen-MM, and MoonDream2 each created $20\%$ of the dataset, while xLAM generated the remaining $40\%$. xLAM is the most accurate model in terms of valid syntax (Table \ref{tbl:miniwob_acc}).
\end{itemize}

\label{app:offline_datasets}

\subsubsection{Additional MiniWoB Results}
The results of Figure \ref{fig:miniwob_main_text} are listed according to the $27$ individual browser tasks in Tables \ref{tbl:miniwob_suc} and \ref{tbl:miniwob_acc}.

\begin{table}[h!]
\centering
\resizebox{.8\linewidth}{!}{

\begin{tabular}{lcccccc}
\toprule
 & PaliGemma & xGenMM & xLAM & MD2 & MD2+SFT & MD2+AFSFT \\
\midrule
click-tab & 0.0 & 3.1 & \textbf{94.8} & 0.0 & 16.7 & \underline{66.5} \\
click-tab-2-easy & 0.0 & \textbf{1.0} & \textbf{1.0} & 0.0 & 0.0 & 0.0 \\
click-dialog-2 & 0.0 & 16.9 & \textbf{84.3} & 0.0 & 21.7 & \underline{66.8} \\
click-shape & 0.0 & 0.8 & \textbf{1.7} & 0.0 & 0.0 & \underline{1.5} \\
click-checkboxes-transfer & 0.0 & 0.0 & \textbf{71.5} & 0.0 & 14.9 & \underline{61.7} \\
click-shades & 0.0 & 0.0 & 0.0 & 0.0 & 0.0 & 0.0 \\
click-checkboxes & 0.0 & 1.1 & \textbf{52.2} & 0.0 & 5.6 & \underline{33.8} \\
click-checkboxes-large & 0.0 & 0.0 & \textbf{5.2} & 0.0 & 0.0 & 0.0 \\
click-tab-2-hard & 0.0 & 0.0 & 0.0 & 0.0 & 0.0 & \textbf{0.5} \\
click-test-transfer & 0.0 & 42.2 & \underline{90.4} & 0.0 & 54.9 & \textbf{91.3} \\
click-collapsible-2 & 0.0 & 0.0 & \textbf{1.0} & 0.0 & 0.0 & 0.0 \\
click-test & 0.0 & 56.8 & \textbf{100.0} & 0.0 & \textbf{100.0} & \textbf{100.0} \\
click-menu & 0.0 & 6.2 & \textbf{31.0} & 0.0 & 3.0 & \underline{23.3} \\
click-collapsible & 0.0 & 0.0 & \textbf{0.3} & 0.0 & 0.0 & 0.0 \\
click-checkboxes-soft & 0.0 & 0.0 & \textbf{33.0} & 0.0 & 3.9 & \underline{29.8} \\
click-button & 0.0 & 10.6 & \textbf{95.6} & 0.0 & 22.8 & \underline{91.7} \\
click-tab-2-medium & 0.0 & 1.0 & \textbf{19.7} & 0.0 & \underline{9.2} & 0.0 \\
click-tab-2 & 0.0 & 0.0 & \textbf{1.9} & 0.0 & 0.0 & 0.0 \\
click-option & 0.0 & 0.0 & \textbf{66.8} & 0.0 & 0.0 & 0.0 \\
click-menu-2 & 0.0 & 0.0 & \textbf{8.9} & 0.0 & 0.0 & 0.0 \\
click-color & 0.0 & 0.0 & 0.0 & 0.0 & 0.0 & 0.0 \\
click-widget & 0.0 & 14.5 & \textbf{60.7} & 0.0 & 16.2 & \underline{58.5} \\
click-dialog & 0.0 & 4.1 & \textbf{100.0} & 0.0 & 32.6 & \textbf{100.0} \\
click-link & 0.0 & 9.3 & 0.0 & 0.0 & 0.5 & \textbf{11.8} \\
click-scroll-list & 0.0 & 1.0 & \textbf{27.4} & 0.0 & 0.0 & 0.0 \\
click-button-sequence & 0.0 & \underline{1.8} & \textbf{5.6} & 0.0 & 0.0 & 0.0 \\
click-test-2 & 0.0 & 45.5 & \textbf{90.4} & 0.0 & 44.6 & \underline{89.2} \\
\bottomrule
\end{tabular}}

\caption{\textbf{BrowserGym MiniWoB Success Rates (\%).} Evaluated over a sample of $5,000$ timesteps per task. \textbf{Bold} and \underline{underlined} entries denote the best and second-best performance per task, respectively.}
\label{tbl:miniwob_suc}
\end{table}

\begin{table}[h!]
\centering
\resizebox{.8\linewidth}{!}{

\begin{tabular}{lcccccc}
\toprule
 & PaliGemma & xGenMM & xLAM & MD2 & MD2+SFT & MD2+AFSFT \\
\midrule
click-tab & 13.3 & 97.5 & \underline{99.7} & 39.8 & 7.0 & \textbf{100.0} \\
click-tab-2-easy & 16.7 & \underline{95.1} & \textbf{99.7} & 49.0 & 6.6 & 91.5 \\
click-dialog-2 & 16.0 & 95.7 & \textbf{100.0} & 44.6 & 6.7 & \underline{99.9} \\
click-shape & 26.6 & \underline{96.1} & \textbf{99.3} & 46.0 & 9.6 & 77.1 \\
click-checkboxes-transfer & 10.4 & 93.4 & \textbf{100.0} & 44.4 & 9.6 & \underline{94.0} \\
click-shades & 28.4 & 91.0 & \underline{99.9} & 47.8 & 74.0 & \textbf{100.0} \\
click-checkboxes & 11.1 & 92.5 & \textbf{100.0} & 42.9 & 10.2 & \underline{95.6} \\
click-checkboxes-large & 4.4 & \underline{95.0} & \textbf{99.5} & 34.3 & 0.0 & 46.0 \\
click-tab-2-hard & 5.8 & \underline{89.8} & \textbf{99.9} & 41.3 & 0.3 & 79.1 \\
click-test-transfer & 33.1 & 96.6 & \textbf{100.0} & 39.0 & 22.7 & \underline{96.8} \\
click-collapsible-2 & 21.5 & 97.6 & \textbf{99.6} & 45.3 & 12.7 & \underline{98.2} \\
click-test & 27.3 & 99.5 & \textbf{100.0} & 37.2 & \textbf{100.0} & \textbf{100.0} \\
click-menu & 22.5 & 92.9 & \textbf{100.0} & 46.6 & 6.4 & \underline{99.8} \\
click-collapsible & 18.2 & 97.5 & \underline{99.9} & 46.7 & 98.9 & \textbf{100.0} \\
click-checkboxes-soft & 7.4 & 96.0 & \textbf{100.0} & 42.5 & 7.0 & \underline{99.5} \\
click-button & 20.0 & \underline{95.3} & \textbf{100.0} & 44.8 & 6.6 & 95.0 \\
click-tab-2-medium & 17.1 & 96.6 & \textbf{99.8} & 48.4 & 5.1 & \underline{97.8} \\
click-tab-2 & 7.8 & \underline{96.4} & \textbf{99.8} & 45.2 & 4.0 & 95.5 \\
click-option & 11.2 & 95.7 & \textbf{100.0} & 44.8 & 9.0 & \underline{96.2} \\
click-menu-2 & 24.4 & \underline{95.9} & \textbf{100.0} & 46.2 & 5.7 & 72.8 \\
click-color & 27.9 & 97.3 & \textbf{99.9} & 41.8 & 30.6 & \underline{99.5} \\
click-widget & 17.4 & 97.0 & \textbf{100.0} & 47.8 & 5.6 & \textbf{100.0} \\
click-dialog & 20.6 & 95.5 & \textbf{100.0} & 43.1 & 7.4 & \textbf{100.0} \\
click-link & 22.4 & \underline{97.3} & \textbf{99.9} & 46.2 & 10.7 & 66.9 \\
click-scroll-list & 8.8 & 97.3 & \textbf{99.9} & 41.6 & 5.9 & \underline{97.8} \\
click-button-sequence & 31.8 & \underline{93.5} & \textbf{99.9} & 40.2 & 24.8 & 80.6 \\
click-test-2 & 29.7 & 97.1 & \textbf{100.0} & 38.1 & 18.7 & \underline{97.3} \\
\bottomrule
\end{tabular}

}
\caption{\textbf{BrowserGym MiniWoB Valid Action Syntax Rates (\%).} \textbf{Bold} and \underline{underlined} entries denote the best and second-best performance per task, respectively.}
\label{tbl:miniwob_acc}
\end{table}

\vspace{50mm}

\subsection{Example Observation Prompts}
\label{app:prompting}

\begin{figure}[htbp]
    \centering
    \begin{minipage}{0.95\textwidth}
\begin{lstlisting}
Actions Taken: 0. Max Actions Allowed: 10.
You are playing a game of Blackjack.
You have long-term memory. If you want to remember something, include '[start remember] MESSAGE [stop remember]' in your response.

You can select an action by writing '[action] CHOICE', where CHOICE is one of the following choices: [ stand , hit ]. The action should be the final part of your answer. For example, a valid action would be '[action] stand'.
<|Text-Specific Info Begins|> You observe: Dealer's card: 9. Player's cards: 8, 8. <|Text-Specific Info Ends|>
Let's think step-by-step. Analyze the current observation, remember key details, and answer this question: Which action do you choose?
\end{lstlisting} % If stored in a file
    \end{minipage}
    \caption{\textbf{Example Prompt from the Gym Cards Blackjack Domain}. \texttt{<|Text-Specific Info|>} markers denote text that is included only for LLM evaluations where the model cannot observe the corresponding image.}
    \label{fig:gym_cards_prompt}
\end{figure}

\begin{figure}[htbp]
    \centering
    \begin{minipage}{0.95\textwidth}
\begin{lstlisting}
Actions Taken: 0. Max Actions Allowed: 64.
You are playing as the red triangle in a gridworld game. The triangle points in the direction you are facing.
You have long-term memory. If you want to remember something, include '[start remember] MESSAGE [stop remember]' in your response.

You select an action with'[action] CHOICE', where CHOICE is one of: [ turn left , turn right , move forward , pick up , drop , toggle , done ]. The action should be the final part of your answer. For example, a valid action would be '[action] turn right'.
Your goal is: go to the yellow box
Look at the image observation. Which action do you choose?
\end{lstlisting} % If stored in a file
    \end{minipage}
    \caption{\textbf{Example prompt from the BabyAI domain.}}
    \label{fig:babyai_prompt}
\end{figure}

\begin{figure}[htbp]
    \centering
    \vspace{-10mm}
    \begin{minipage}{\textwidth}
\begin{lstlisting}
Actions Taken: 0. Max Actions Allowed: 25.
You are a browser agent. Your goal is: Switch between the tabs to find and click on the link "neque".
You have long-term memory. If you want to remember something, include '[start remember] MESSAGE [stop remember]' in your response.

Select an Action by writing '[action] `ACTION`' as the final part of your answer. Your `ACTION` should be a short piece of python code in the following format: 
12 different types of actions are available.

noop(wait_ms: float = 1000)
    Examples:
        noop()

        noop(500)

scroll(delta_x: float, delta_y: float)
    Examples:
        scroll(0, 200)

        scroll(-50.2, -100.5)

fill(bid: str, value: str)
    Examples:
        fill('237', 'example value')

        fill('45', 'multi-line\nexample')

        fill('a12', 'example with "quotes"')

select_option(bid: str, options: str | list[str])
    Examples:
        select_option('a48', 'blue')

        select_option('c48', ['red', 'green', 'blue'])

click(bid: str, button: Literal['left', 'middle', 'right'] = 'left', modifiers: list[typing.Literal['Alt', 'Control', 'ControlOrMeta', 'Meta', 'Shift']] = [])
    Examples:
        click('a51')

        click('b22', button='right')

        click('48', button='middle', modifiers=['Shift'])

dblclick(bid: str, button: Literal['left', 'middle', 'right'] = 'left', modifiers: list[typing.Literal['Alt', 'Control', 'ControlOrMeta', 'Meta', 'Shift']] = [])
    Examples:
        dblclick('12')

        dblclick('ca42', button='right')

        dblclick('178', button='middle', modifiers=['Shift'])

hover(bid: str)
    Examples:
        hover('b8')

press(bid: str, key_comb: str)
    Examples:
        press('88', 'Backspace')

        press('a26', 'ControlOrMeta+a')

\end{lstlisting}
    \end{minipage}
    \vspace{-2mm}
    \caption{\textbf{Example prompt from the MiniWoB domain}. The goal description varies by task. \textbf{(Part 1/2)}.}
    \label{fig:miniwob_prompt_pt1}
\end{figure}

\begin{figure}[htbp]
    \centering
    \vspace{-10mm}
    \begin{minipage}{\textwidth}
\begin{lstlisting}
focus(bid: str)
    Examples:
        focus('b455')

clear(bid: str)
    Examples:
        clear('996')

drag_and_drop(from_bid: str, to_bid: str)
    Examples:
        drag_and_drop('56', '498')

upload_file(bid: str, file: str | list[str])
    Examples:
        upload_file('572', 'my_receipt.pdf')

        upload_file('63', ['/home/bob/Documents/image.jpg', '/home/bob/Documents/file.zip'])

Only a single action can be provided at once. Example:
fill('a12', 'example with "quotes"')
 An example of a complete correct output is '[action] click('48', button='middle', modifiers=['Shift'])'.


 Accessibility Tree ([bid] info, center=(x coord, y coord)): 
 RootWebArea 'Click Tab Task', focused, url='file:///export/home/miniwob-plusplus/miniwob/html/miniwob/click-tab-2.html'
        [17] tablist '', center="(80,68)", multiselectable=False, orientation='horizontal'
                [18] tab 'Tab #1', center="(27,70)", expanded=True, selected=True, controls='tabs-1'
                        [19] link 'Tab #1', center="(27,70)", url='file:///export/home/miniwob-plusplus/miniwob/html/miniwob/click-tab-2.html#tabs-1'
                [20] tab 'Tab #2', center="(71,69)", expanded=False, selected=False
                        [21] link 'Tab #2', center="(71,70)", url='file:///export/home/miniwob-plusplus/miniwob/html/miniwob/click-tab-2.html#tabs-2'
                [22] tab 'Tab #3', center="(115,69)", expanded=False, selected=False
                        [23] link 'Tab #3', center="(115,70)", url='file:///export/home/miniwob-plusplus/miniwob/html/miniwob/click-tab-2.html#tabs-3'
        [24] tabpanel 'Tab #1', center="(80,137)"
                [25] paragraph '', center="(80,137)"
                        StaticText 'Enim, aliquet risus pellentesque commodo'
                        StaticText 'in'
                        StaticText 'nibh. Venenatis'
                        StaticText 'id.'
                        StaticText 'Adipiscing. Feugiat justo tellus. Tortor cum'
                        StaticText 'convallis'
                        StaticText 'dolor quisque id egestas.'

<|Image-Specific Info Begins|>
Web elements in the image have been labeled with their `bid` values.
<|Image-Specific Info Ends|>
You previously output: '' with error message: None

Determine the next action to accomplish your goal. Make sure to follow the formatting instructions. Which action should you take?
\end{lstlisting}
    \end{minipage}
    \caption{\textbf{Example prompt from the MiniWoB domain.} The goal description varies by task. \texttt{<|Image-Specific Info|>} markers denote text that is included only for VLM evaluations where the agent can observe the corresponding browser screenshot. \textbf{(Part 2/2)}.}
    \label{fig:miniwob_prompt_pt2}
\end{figure}

\end{document}